\documentclass[conference]{IEEEtran}

\usepackage{times}
\usepackage{epsfig}
\usepackage{graphicx}
\usepackage{amsmath}
\usepackage{amssymb}

\usepackage{color}
\usepackage{bm}

\usepackage[caption=false,font=footnotesize]{subfig}

\usepackage[capitalise]{cleveref}
\usepackage{booktabs}
\usepackage{multirow}
\usepackage{cite}
\usepackage{verbatim} 

\begin{document}

\title{Optimize Deep Convolutional Neural Network with Ternarized Weights and High Accuracy}



\author{\IEEEauthorblockN{Zhezhi He\IEEEauthorrefmark{1}, Boqing Gong\IEEEauthorrefmark{2}, and
Deliang Fan\IEEEauthorrefmark{1}}

\IEEEauthorblockA{\IEEEauthorrefmark{1}Department of Electrical and Computer Engineering,
University of Central Florida, Orlando, 32816 USA \\
\IEEEauthorrefmark{3} Tencent AI lab, Seattle, USA
}
\textit{(Elliot.He@knighst.ucf.edu; Dfan@ucf.edu)}
}

\maketitle

\begin{abstract}
Deep convolution neural network has achieved great success in many artificial intelligence applications. However, its enormous model size and massive computation cost have become the main obstacle for deployment of such powerful algorithm in the low power and resource limited embedded systems. As the countermeasure to this problem, in this work, we propose statistical weight scaling and residual expansion methods to reduce the bit-width of the whole network weight parameters to ternary values (i.e. -1, 0, +1), with the objectives to greatly reduce model size, computation cost and accuracy degradation caused by the model compression. With about 16$\times$ model compression rate, our ternarized ResNet-32/44/56 could outperforms full-precision counterparts by 0.12\%, 0.24\% and 0.18\% on CIFAR-10 dataset. We also test our ternarization method with AlexNet and ResNet-18 on ImageNet dataset, which both achieve the best top-1 accuracy compared to recent similar works, with the same 16$\times$ compression rate. If further incorporating our residual expansion method, compared to the full-precision counterpart, our ternarized ResNet-18 even improves the \textbf{top-5 accuracy by 0.61\%} and merely degrades the top-1 accuracy only by 0.42\% for ImageNet dataset, with 8$\times$ model compression rate. It outperforms the recent ABC-Net by 1.03\% in top-1 accuracy and 1.78\% in top-5 accuracy, with around 1.25$\times$ higher compression rate and more than 6$\times$ computation reduction due to the weight sparsity.   

\end{abstract}

\section{Introduction}

Deep convolutional neural networks (CNNs) have taken an important role in artificial intelligence algorithm which has been widely used in computer vision, speech recognition, data analysis and etc \cite{lecun2015deep}. Nowadays, deep CNNs become more and more complex consisting of more layers, larger model size and denser connections. However, from the hardware acceleration point of view, deep CNNs still suffer from the obstacle of hardware deployment due to their massive cost in both computation and storage. For instance, VGG-16 \cite{sung2015resiliency} from ILSVRC 2014 requires 552MB of parameters and 30.8 GFLOP per image, which is difficult to deploy in resource limited mobile systems. Research has shown that deep CNN contains significant redundancy, and the state-of-the-art accuracy can also be achieved through model compression \cite{han2015deep}. Many recent works have been proposed to address such high computational complexity and storage capacity issues of existing deep CNN structure using model compression techniques. 

As one of the most popular model compression technique, weight quantization techniques are widely explored in many related works \cite{han2015deep, zhu2016trained, zhou2016dorefa,tang2017train,rastegari2016xnor} which can significantly shrink the model size and reduce the computation complexity. It is worthy to note that weight binarization (-1, +1) or ternarization (-1, 0, +1) is more preferred compared to other methods since floating point multiplication is not needed and most complex convolution operations are converted to bit-wise \textit{xnor} and \textit{bit-count} operations, which could greatly save power, area and latency. Meanwhile, it does not modify the network topology or bring extra computation to the system, which may increase hardware deployment complexity, such as pruning or hash compression. However, for compact deep CNN models that are more favored by hardware deployment, the existing aggressive weight binarization or ternarization methods still encounter large accuracy degradation for large scale benchmarks. 

In this work, we propose statistical weight scaling and residual expansion methods to convert the entire network weight parameters to ternary format (i.e. -1, 0, +1), with objectives to greatly reduce model size, computation cost and accuracy degradation caused by model compression. Our main contributions in this work can be summarized as:

\begin{itemize}
    \item A iterative statistical weight ternarization method is used to optimize the layer-wise scaling factor to mitigate accuracy degradation caused by aggressive weight ternarization. Such method leads to that the ternarization of ResNet32/44/56 even outperforms full-precision counterparts by 0.12\%, 0.24\% and 0.18\% for CIFAR-10 dataset. Moreover, both the ternarization of AlexNet and ResNet-18 in ImageNet dataset achieve the best top-1 accuracy compared to recent similar works, with the same 16X compression rate.

    \item Different from other existing works that leave the network's first and last layers in full-precision, we ternarize all the convolution and fully-connected layers, where a very limited accuracy degradation is observed. Such whole network ternarization could bring large amount of power, area and latency saving for domain-specific deep CNN accelerators. 
    
    \item To further mitigate accuracy degradation from the full-precision baseline, we introduce a residual expansion methodology. Compared to full-precision network, our ternarized ResNet-18 even improves top-5 accuracy by 0.61\% and degrades top-1 accuracy only by 0.42\% for ImageNet dataset, with 8$\times$ model compression rate. It outperforms the recent ABC-Net by 1.03\% in top-1 accuracy and 1.78\% in top-5 accuracy, with 1.25$\times$ higher compression rate.
    
\end{itemize}

\section{related works}
\label{sec_related_works}
Recently, model compression on deep convolutional neural network has emerged as one hot topic in the hardware deployment of artificial intelligence. As the most popular technique, weight quantization techniques are widely explored in many related works which can significantly shrink the model size and reduce the computation complexity. Among all those works, DCNN with binary weight is the most discussed scheme, since it leads to 32x model compression rate. More importantly, it also converts the original floating-point multiplication (i.e. \textit{mul}) operations into addition/subtraction (i.e. \textit{add/sub}), which could greatly reduce computational hardware resources and further dramatically lowers the existing barrier to deploy powerful deep neural network algorithm into low power resource-limited embedded system. 
BinaryConnect \cite{courbariaux2015binaryconnect} is the first work of binary CNN which can get close to the state-of-the-art accuracy on CIFAR-10, whose most effective technique is to introduce the gradient clipping. After that, both BWN in \cite{rastegari2016xnor} and DoreFa-Net \cite{zhou2016dorefa} show better or close validation accuracy on ImageNet dataset. In order to reduce the computation complexity to the bone, XNOR-Net \cite{rastegari2016xnor} binarize the input tensor of the convolution layer which further converts the \textit{add/sub} operations into bit-wise \textit{xnor} and \textit{bit-count} operations. Besides weight binarization, there are also recent works proposing to ternarize the weights of neural network using trained scaling factors \cite{zhu2016trained}. Leng et. al. employ ADMM method to optimize neural network weights in configurable discrete levels to trade off between accuracy and model size \cite{leng2017extremely}. ABC-Net in \cite{lin2017towards} proposes multiple parallel binary convolution layers to improve the network model capacity and accuracy, while maintaining binary kernel. All above discussed aggressive neural network binarization or ternarization methodologies sacrifice the inference accuracy in comparison with the full precision counterpart to achieve large model compression rate and computation cost reduction.

\section{Weight Ternarization Training Method}
\label{sec_ternarization}
\subsection{Ternarization with iterative statistical scaling}

\begin{figure*}[t]
	\centering
		\begin{tabular}{l}
		\includegraphics [width=0.75\linewidth]{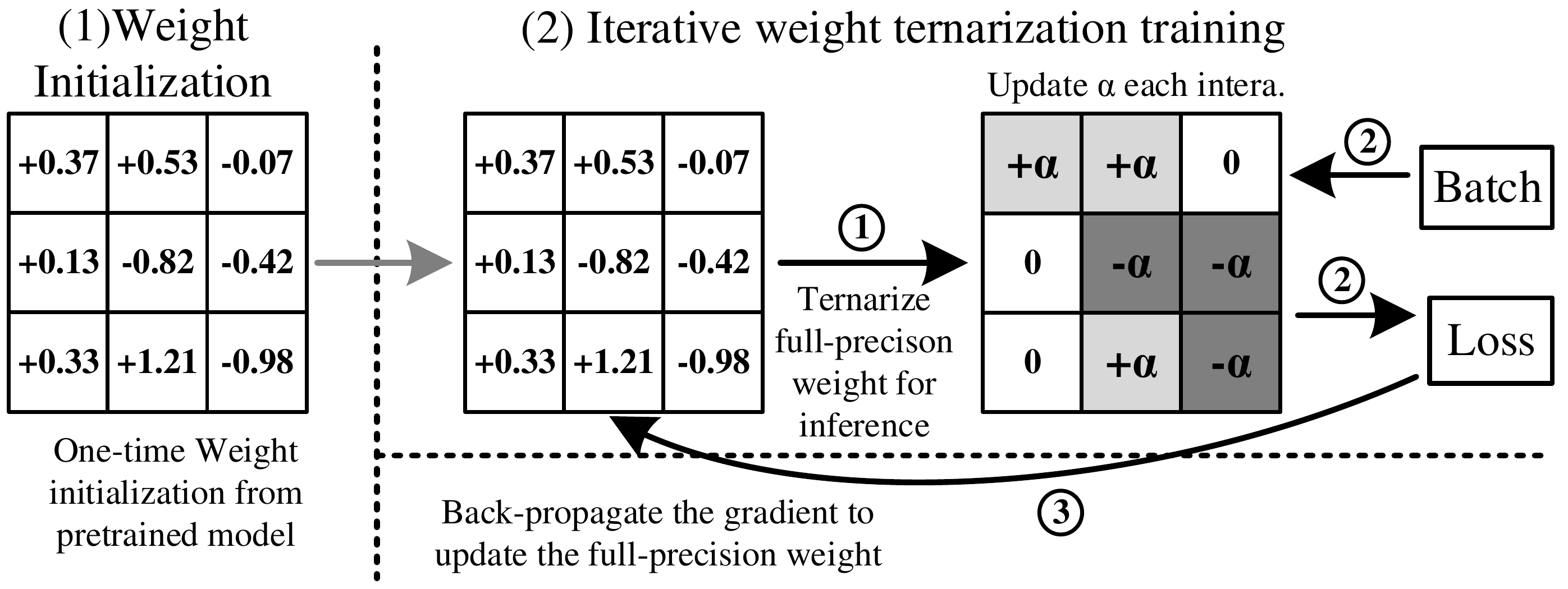}\\
		\end{tabular}
	\caption{The training scheme of ternarized CNN model. \textcircled{\raisebox{-0.1pt}{1}}-\textcircled{\raisebox{-0.1pt}{3}} are iteratively operated during the training.}
	\label{fig_train_flow}
\end{figure*}
As shown in \cref{fig_train_flow}, our training methodology to obtain accurate deep Convolutional Neural Network (CNN) 
\footnote{In this work, all the convolution kernels and fully-connected layers do not have bias term, excluding the last layer. } with ternarized weights can be divided into two main steps: Initialization and iterative weight ternarization training. We first train a designated neural network model with full precision weights to act as the initial model for the future ternarized neural network model. After model initialization, we retrain the ternarized model with iterative inference (including \textit{statistical weight scaling}, weight ternarization and ternary weight based inference for computing loss function) and back-propagation to update full precision weights.

\textbf{Initialization:} The reason that we use a pretrained model with full precision weights as the initial model comes in twofold: 1) Fine-tuning from the pretrained model normally gives higher accuracy for the quantized neural network. 2) Our ternarized model with the initialized weights from pretrained model converges faster in comparison to ternarization training from scratch. For practical application, engineers will definitely use a full-precision model to estimate the ideal upper-bound accuracy. Afterwards, various model compression techniques are applied to the full-precision model to compress model size while trying to mitigate accuracy loss as well.

\textbf{Iterative weight ternarization training:}
After loading the full-precision weights as initialization state, we start to fine-tune the ternarized CNN model. 
As described in \cref{fig_train_flow}, there are three iterative steps for the following training, namely, \textcircled{\raisebox{-0.9pt}{1}}-statistical weight scaling and weight ternarization, \textcircled{\raisebox{-0.9pt}{2}}-ternary weight based inference for loss function computation and \textcircled{\raisebox{-0.9pt}{3}}-back propagation to update full precision weights.

\textcircled{\raisebox{-0.9pt}{1}} will first ternarize current full precision weights and compute the corresponding scaling factor based on current statistical distribution of full precision weight. For weight ternarization (i.e. -1, 0, +1), we adopt the variant staircase ternarization function, which compares the full precision weight with the symmetric threshold $\pm\Delta_{th}$ \cite{li2016ternary}. Such weight ternarization function in the forward path for inference can be mathematically described as:
\begin{equation}
\label{eqt_sign2}
    \textup{Forward}:~~~ \bm{w}_l' = 
    \begin{cases}
    \alpha\times Sign(w_{l,i}) & |w_{l,i}|\geq \Delta_{th}\\
    0         & |w_{l,i}| < \Delta_{th}\\
    \end{cases}
\end{equation}
where $\bm{w}_l$ denotes the full precision weight tensor of layer $l$, $\bm{w'}_l$ is the weight after ternarization. We set the scaling factor as:
\begin{equation}
    \alpha = E(|\bm{w}_{l,i}|), ~~~ \forall \{i\big\vert|\bm{w}_{l,i}|\geq\Delta_{th}\}
\end{equation}  
which statistically calculates the mean of absolute value of designated layer's full precision weights that are greater than the threshold $\Delta_{th}$. Unlike TWN uses $\Delta_{th}=0.7\times E(\bm{w}_{l})$ as threshold, we set $\Delta_{th}=\beta\times max(|\bm{w}_{l}|)$ as threshold \cite{li2016ternary, zhu2016trained}. The reason is that, for each training iteration, the weight update causes large value drifting of $E(\bm{w}_{l})$, which correspondingly leads to unstable $\Delta_{th}$. In this work, we employ single scaling factor for both positive and negative weights, since such symmetric weight scaling factor can be easily extracted and integrated into following Batch Normalization layer or ReLU activation function (i.e., both perform element-wise function on input tensor) for the hardware implementation. 

Then, in \textcircled{\raisebox{-0.9pt}{2}}, the input mini-batch takes the ternarized model for inference and calculates loss w.r.t targets. In this step, since all of the weights are in ternary values, all the dot-product operations in layer $l$ can be expressed as:
\begin{equation}
     \bm{x}_{l}^T\cdot \bm{w}_{l}' = \bm{x}_{l}^T\cdot (\alpha \cdot Tern(\bm{w}_l)) = \alpha \cdot( \bm{x}_{l}^T \cdot Tern(\bm{w}_l) )
\end{equation}
where $\bm{x}_{l}$ is the vectorized input of layer $l$. Since $Tern(\bm{w}_{l,i})\in\{-1,0,+1\}$, $\bm{x}_{l}^T \cdot Tern(\bm{w}_l)$ can be easily realized through addition/subtraction without multi-bit or floating point multiplier in the hardware, which greatly reduces the computational complexity.

In \textcircled{\raisebox{-0.9pt}{3}}, the full precision weights will be updated during back-propagation. Then, the next iteration begins to re-compute weight scaling factor and ternarize weights as described in step-\textcircled{\raisebox{-0.9pt}{1}}. Meanwhile, since the ternarization function owns zero derivatives almost everywhere, which makes it impossible to calculate the gradient using chain rule in backward path. Thus, the Straight-Through Estimator (STE) \cite{bengio2013estimating}\cite{zhou2016dorefa} of such ternarization function in the back-propagation is applied to calculate the gradient as follow:
\begin{equation}
    \textup{Backward}:~~~ \frac{\partial g}{\partial w}=
    \begin{cases}
    \frac{\partial g}{\partial w'} & if~|w|\leq 1\\
    0          & otherwise
    \end{cases}
\end{equation}
where the gradient clipping prevents the full precision weight growing too large and stabilize the threshold $\beta\times max(|\bm{w}_{l}|)$ during training (i.e., fine-tuning), thus leading to higher inference accuracy ultimately. 

\subsection{Residual expansion to improve accuracy}

\begin{figure*}[h]
	\centering
		\begin{tabular}{l}
		\includegraphics [width=0.85\linewidth]{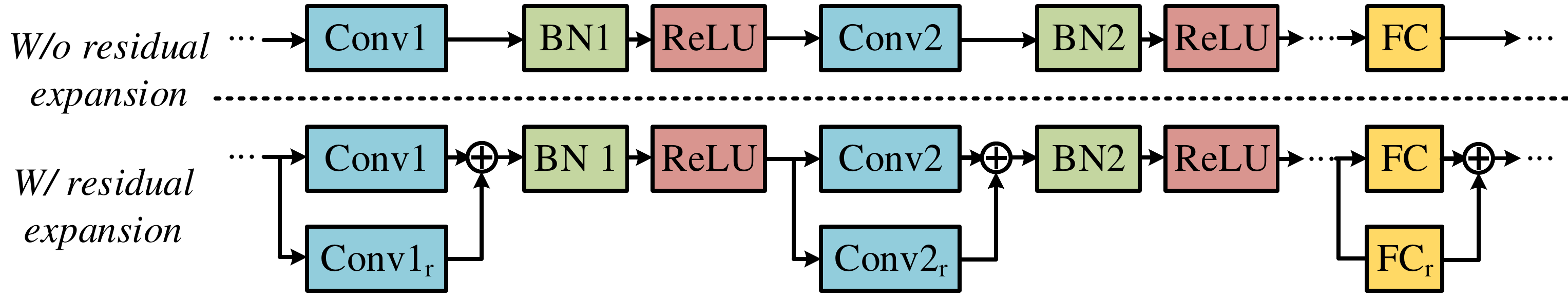}\\
		\end{tabular}
	\caption{The block diagram of (top) original network topology and (bottom) the network with residual expansion to further compensate the accuracy degradation. In this figure, the expansion factor $T_{ex}$ is 2.}
	\label{fig_residual_expansion}
\end{figure*}

As the aforementioned discussion in the introduction, works like DoreFa-net \cite{han2015deep, zhou2016dorefa} adopt the multilevel quantization scheme to preserve the model capacity and avoid the accuracy degradation caused by extremely aggressive (i.e. binary/ternary) weight quantization.
Since our main objective is to totally remove floating point multiplication and only use addition/subtraction to implement the whole deep CNN, while minimizing accuracy degradation caused by weight ternarization. We propose a \textit{Residual Expanded Layer} (REL) as the remediation method to reduce accuracy degradation. As depicted in \cref{fig_residual_expansion}, we append one REL over the convolution layers and fully-connected layers in the original network structure, where the expansion factor $T_{ex}$ is 2 in this case. We use \textit{Conv} and \textit{Conv\textsubscript{r}} to denote the original layer and the added REL respectively.
Both \textit{Conv} and \textit{Conv\textsubscript{r}} are ternarized from the identical full precision parameters. During training, the network structure with RELs follow the same procedures as discussed in \cref{fig_train_flow}. The only difference is that \textit{Conv} and \textit{Conv\textsubscript{r}} has two different threshold factors (e.g., $\beta=\{0.05, 0.1\}$ are in this work for $T_{ex}=2$). We could append more RELs for compact CNN with low network redundancy.


The appended REL equivalently introduces more number of levels to each kernel. Assuming two threshold factors $a$, $b$ for \textit{Conv} and \textit{Conv\textsubscript{r}} layers, where $a>b$, then we can formulate their computation as: 
\begin{equation}
    \bm{x}^T\cdot \bm{w}' + \bm{x}^T\cdot \bm{w}'_{r}=\bm{x}^T\cdot (\alpha\cdot \underset{\beta=a}{Tern}(\bm{w})+\alpha_r\cdot \underset{\beta=b}{Tern}(\bm{w}_r)) 
\end{equation}
where the function $\alpha\cdot\underset{\beta=a}{Tern}(\bm{w})+\alpha_r\cdot \underset{\beta=b}{Tern}(\bm{w}_r)$ is equivalent to a multi-threshold function that divides the element in $\bm{w}$ into levels of $\gamma = \{-\alpha-\alpha_r, -\alpha, 0, \alpha, \alpha+\alpha_r\},~\gamma_i\in[2\times E(|\bm{w}|), ~2\times max(\bm{w})]$, with thresholds $\{-b, -a, 0, a, b\}$. Enlarging the expansion factor $T_{ex}$ will equivalently introduce more number of levels for the weights and recover the ternarized model capacity towards its full-precision baseline. However, compared to the case that directly use a multi-bit kernels, the actual convolution kernel is still in ternary format and thus no multi-bit multiplication is needed (see more discussions in \cref{sec_multibit}).

\section{Experiments}

In this work, all experiments are performed under the framework of Pytorch. The performance of our neural network ternarization methodology is examined using two datasets: CIFAR-10 and ImageNet. Other small datasets like MNIST and SVHN are not studied here since other related work \cite{zhou2016dorefa, li2016ternary} can already obtain close or even no accuracy loss.

\subsection{CIFAR-10}
To impartially compare our ternarization method with \cite{zhu2016trained}, we choose the same ResNet-20, 32, 44, 50 (type-A parameter free residual connection) \cite{he2016deep} as the testing neural network architecture on CIFAR-10 dataset. CIFAR-10 contains 50 thousands training samples and 10 thousands test samples with $32\times 32$ image size. The data augmentation method is identical as used in \cite{he2016deep}. For fine-tuning, we set the initial learning rate as 0.1, which is scheduled to scale by 0.1 at epoch 80, 120 and 160 respectively. Note that, both the first and last layer are ternarized during the training and test stage. 

we report the CIFAR-10 inference accuracy as listed in \cref{table_cifar10}. It can be seen that only ternarized ResNet-20 gets minor (0.64\%) accuracy degradation, while all the other deeper networks like ResNet-32, 44, 56 actually outperform the full precision baseline with 0.12\%, 0.24\% and 0.18\%, respectively. It shows that a more compact neural network, like ResNet20, is more likely to encounter accuracy loss, owing to the network capacity is hampered by this aggressive model compression. Similar trend is also reported in TTN \cite{zhu2016trained}. Meanwhile, except ResNet56, other accuracy improvements of our method are higher. Note that, the improvement is computed as the accuracy gap between pre-trained full precision model and corresponding ternarized model. Our pre-trained full precision model accuracy is slightly different even with the same configurations. 

\begin{table}[h]
\centering
\caption{Inference accuracy of ResNet on CIFAR-10 dataset}
\label{table_cifar10}
\scalebox{1}{
\begin{tabular}{@{}cccccc@{}}
\toprule
 &  & \footnotesize{ResNet20} & \footnotesize{ResNet32} & \footnotesize{ResNet44} & \footnotesize{ResNet56} \\ \midrule
\multirow{2}{*}{TTN \cite{zhu2016trained}} & FP & 91.77 & 92.33 & 92.82 & 93.2 \\
 & Tern & 91.13 & 92.37 & 92.98 & 93.56 \\ 
 & Gap  & -0.64  & +0.04  & +0.16  & +0.36  \\ \hline
\multirow{3}{*}{our work} & FP & 91.7 & 92.36 & 92.47 & 92.68\\
 & Tern & 91.65 & 92.48 & 92.71 & 92.86 \\ 
 & Gap  & -0.05  & +0.12  & +0.24  & +0.18  \\
\bottomrule 
\end{tabular}}
\end{table}

\subsection{ImageNet}

In order to provide a more comprehensive experimental results on large dataset, we examine our model ternarization techniques on image classification task with ImageNet \cite{russakovsky2015imagenet} (ILSVRC2012) dataset. ImageNet contains 1.2 million training images and 50 thousands validation images, which are labeled with 1000 categories. For the data pre-processing, we choose the scheme adopted by ResNet \cite{he2016deep}. Augmentations applied to the training images can be sequentially enumerated as: $224\times 224$ randomly resized crop, random horizontal flip, pixel-wise normalization. All the reported classification accuracy on validation dataset is single-crop result. Beyond that, similar as other popular model quantization works, such as XNOR-Net \cite{rastegari2016xnor}, DoreFa-Net \cite{zhou2016dorefa} and TTN \cite{zhu2016trained}, we first keep the first and last layer in full precision (i.e. 32-bit float) and ternarize the remaining layers for fair comparison. In addition of that, we also conduct experiments to fully ternarize the whole network. Thus in this case, the weight parameters of the whole network is in ternary format.

\begin{figure}[t]
\centering
	\subfloat[\label{fig_accuracy_alexnet_ff_lf} first\&last layer FP]{%
		\includegraphics[width=0.23\textwidth]{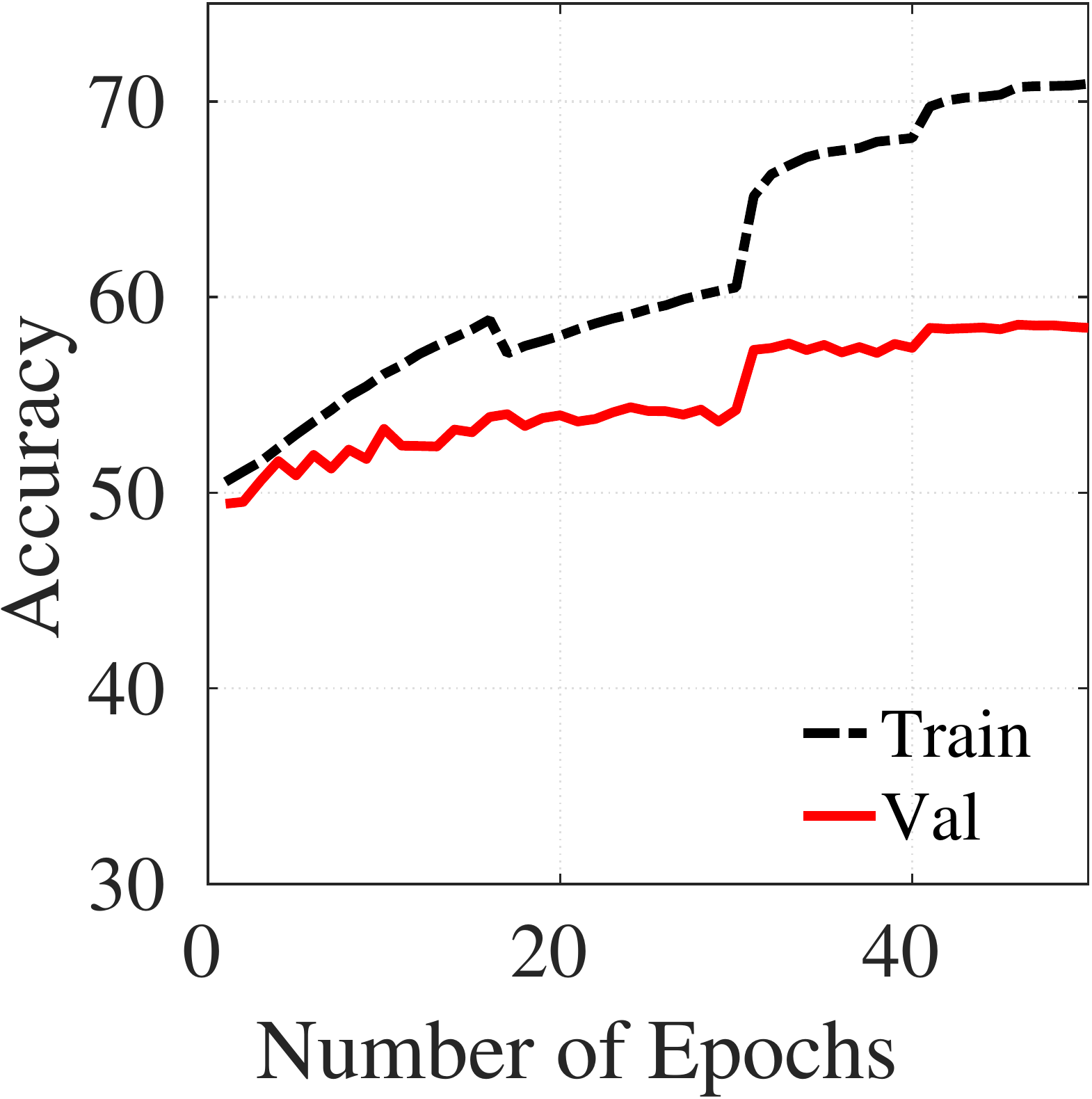}
	} \hspace*{\fill}
	\subfloat[\label{fig_accuracy_alexnet_fq_lq}first\&last layer Tern.]{%
		\includegraphics[width=0.23\textwidth]{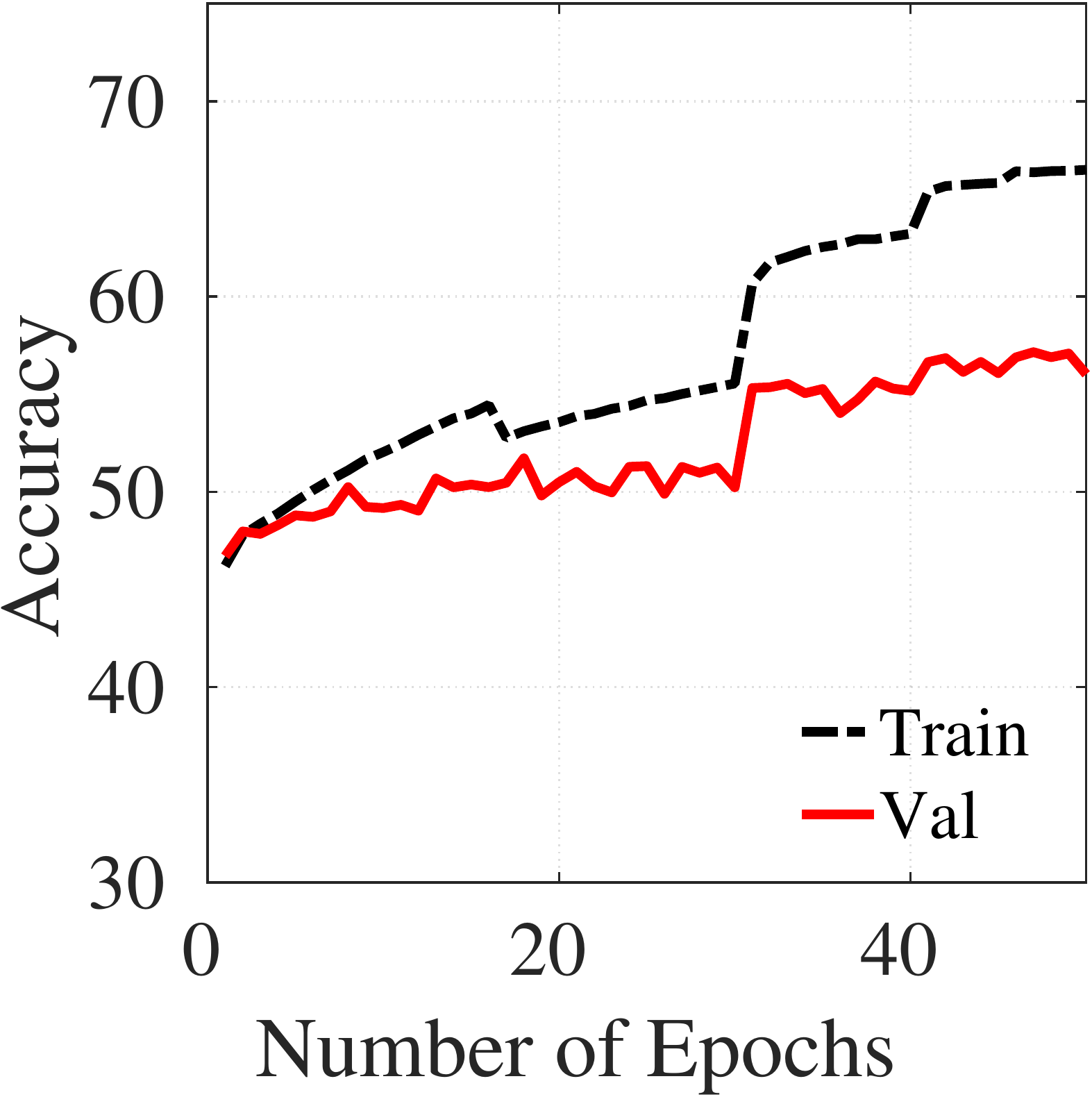}
	}	
	\caption{Train and validation (top-1) accuracy of AlexNet on ImageNet} 
	\label{fig_accuracy_alexnet}
\end{figure}

\subsubsection{AlexNet:}
\label{sec_alexnet}
We first present the experimental results and its analysis on AlexNet, which is similar as the variant AlexNet model used by DoreFa-Net \cite{zhou2016dorefa} and TTN \cite{zhu2016trained}. The AlexNet in this work does not include Local-Response-Normalization (LRN) \cite{krizhevsky2012imagenet}. 
We apply a batch normalization layer after each convolution and fully-connected layer. The dropout layer is removed, since we find that it actually hampers the performance of fine-tuning of ternarized model from its full precision counterpart. A possible reason is that dropout method randomly deactivates some weighted connections. However our kernel scaling factor is calculated with respect to the weight distribution of entire layer. 

We first train the full precision AlexNet with the aforementioned data augmentation method. SGD with momentum = 0.9 is taken as the optimizer. The initial learning rate is 0.1 which is scheduled with decay rate of 0.1 at epoch 45 and 65. Batch size is set to 256, and weight decay is 1e-4. For fine-tuning ternarized AlexNet from the pre-trained full-precision model, we alter the optimizer to Adam. The initial learning rate is set to 1e-4 and scaled by 0.2, 0.2 and 0.5 at epoch 30, 40 and 45 respectively. Batch size is 256 as the full-precision model. Since low-bit quantization could also be considered as a strong regularization techniques, weight decay is expected to be at small value or even zero. Here we set the weight decay as 5e-6.

\begin{table}[h]
\centering
\caption{Validation accuracy of AlexNet on ImageNet using various model quantization methods. FP, Bin., Tern. denote full-precision, binary and ternary respectively.} 
\label{table_alexnet_imagenet}
\scalebox{1}{
\begin{tabular}{@{}cccccc@{}}
\toprule
     & 
\begin{tabular}[c]{@{}c@{}}Quantize\\scheme\end{tabular}&
\begin{tabular}[c]{@{}c@{}}First\\layer \end{tabular} & 
\begin{tabular}[c]{@{}c@{}}Last\\layer\end{tabular} & 
\begin{tabular}[c]{@{}c@{}}Accuracy\\ (top1/top5)\end{tabular} & \begin{tabular}[c]{@{}c@{}}Comp.\\ rate\end{tabular} \\ \midrule
DoreFa-Net\cite{zhou2016dorefa, zhu2016trained} & Bin. & FP & FP & 53.9/76.3 & $\sim$32$\times$ \\
BWN\cite{rastegari2016xnor} & Bin. & FP & FP & 56.8/79.4  & $\sim$32$\times$ \\
ADMM\cite{leng2017extremely}& Bin. & FP* & FP* & 57.0/79.7 & $\sim$32$\times$ \\ \midrule
TWN\cite{li2016ternary,leng2017extremely} & Tern. &  FP & FP  & 57.5/79.8  & $\sim$16$\times$ \\
ADMM\cite{leng2017extremely}& Tern. & FP* & FP* & 58.2/80.6  & $\sim$16$\times$ \\
TTN\cite{zhu2016trained}& Tern. &  FP & FP  & 57.5/79.7  & $\sim$16$\times$ \\
\midrule
Full precision & - & FP & FP & 61.78/82.87  & 1$\times$ \\
\midrule
this work & Tern. & FP & FP & 58.59/80.44 & $\sim$16$\times$ \\
this work & Tern. & Tern & Tern & 57.15/79.42  &  $\sim$16$\times$ \\\bottomrule
\end{tabular}}

\end{table} 

We here report the accuracy on ImageNet validation set including our work and other recently published works with the state-of-the-art performance in \cref{table_alexnet_imagenet}. Note that, we mark out the quantization state of the first layer and last layer for various methods. For works without such discussion in the paper and no released code, we consider that they configure the first and last layer in full precision (marked with *). We are confident in such assumption since we double checked the works they compared in their published papers are with first- and last-layer in full precision. The results show that our work achieves the best accuracy compared to most recent works. In comparison to the previous best result reported by ADMM \cite{leng2017extremely}, our work improves the ADMM binary scheme and ternary scheme by 1.59\% and 0.39\% respectively.

Furthermore, different from previous works that all preserve the weights of first and last layer in full-precision, we also conduct experiments to fully ternarize the whole network. Such ternarization leads to a further 1.44\% top-1 accuracy drop. For a general purpose processor, like CPU/GPU, the performance improvement by ternarizing the whole network is not quite huge since the parameter size reduction and computation cost reduction are relatively small. However, for a domain-specific neural network accelerator, like ASIC and FPGA, ternarizing the entire network means no specific convolution cores are needed to perform Multiplication and Accumulation (MAC) operations \cite{chen201614}. It will save large amount of power, area and other on-chip hardware resources, which further improve the system performance. We will provide more quantitative analysis about this issue in the later discussion section.

\subsubsection{ResNet:} 
\label{sec_resnet}
Various works \cite{zhu2016trained,tang2017train} have drawn a similar conclusion that, under the circumstance of using identical model compression technique, accuracy degradation of the compressed model w.r.t its full-precision baseline will be enlarged when the model topology becomes more compact. In other words, neural network with smaller model size is more likely to encounter accuracy degradation in weight compression. Meanwhile, since neural network with residual connections is the main stream of deep neural network structure, we also employ the compact ResNet-18 (type-b residual connection in \cite{he2016deep}) as another benchmark to demonstrate that our ternarization method could achieves the best state-of-the-art performance. The full-precision ResNet18 model \footnote{https://github.com/pytorch/vision/blob/master/torchvision/models/resnet.py} released by pytorch framework is used as the baseline and pretrained model in this work. The data augmentation method is same as we introduced in \cref{sec_alexnet} for AlexNet.

\begin{table}[h]
\centering
\caption{Validation accuracy (top1/top5 \%) of ResNet-18b \cite{he2016deep} on ImageNet using various model quantization methods.}
\label{table_resnet_imagenet}
\scalebox{1}{
\begin{tabular}{@{}cccccc@{}}
\toprule
     & 
\begin{tabular}[c]{@{}c@{}}Quan.\\scheme\end{tabular}&
\begin{tabular}[c]{@{}c@{}}First\\layer \end{tabular} & 
\begin{tabular}[c]{@{}c@{}}Last\\layer\end{tabular} & 
\begin{tabular}[c]{@{}c@{}}Accuracy\\ (top1/top5)\end{tabular}  & \begin{tabular}[c]{@{}c@{}}Comp.\\ rate\end{tabular} \\ \midrule
BWN\cite{rastegari2016xnor} & Bin. & FP & FP & 60.8/83.0  & $\sim$32$\times$ \\
ABC-Net\cite{lin2017towards} & Bin. & FP* & FP* & 68.3/87.9  &  $\sim$6.4$\times$\\
ADMM\cite{leng2017extremely}& Bin. & FP* & FP* & 64.8/86.2  & $\sim$32$\times$ \\
\midrule
TWN\cite{li2016ternary,leng2017extremely} & Tern. &  FP & FP  & 61.8/84.2  & $\sim$16$\times$ \\
TTN\cite{zhu2016trained}& Tern. &  FP & FP  & 66.6/87.2  & $\sim$16$\times$ \\
ADMM\cite{leng2017extremely}& Tern. & FP* & FP* & 67.0/87.5  & $\sim$16$\times$ \\
\midrule
Full precision & - & FP & FP & 69.75/89.07 & 1$\times$ \\
\midrule
this work & Tern. & FP & FP & 67.95/88.0 & $\sim$16$\times$ \\
this work & Tern. & Tern & Tern & 66.01/86.78 & $\sim$16$\times$ \\ \bottomrule
\end{tabular}}

\end{table}

As the simulation results listed in \cref{table_resnet_imagenet}, ResNet-18b \cite{he2016deep} with our ternarization scheme shows 0.95\% and 0.5\% higher top-1 and top-5 accuracy, respectively, compared to previous best result of ADMM \cite{leng2017extremely} with equal compression rate (16$\times$). ABC-Net \cite{lin2017towards} has $\sim0.3\%$ higher top-1 accuracy over our work, but with a cost of 2.5 times larger model size. Moreover, our work save $\sim$60\% computation owing to the sparse kernel after ternarization. Similar as in AlexNet, if we further ternarize the first and last layer, both the top-1 and top-5 accuracy of ResNet-18 degrades.

\begin{table}[h]
\centering
\caption{Inference accuracy (Top1/Top5 \%) of ternarized ResNets on ImageNet dataset}
\label{table_deepresnet_imagenet}
\scalebox{1}{
\begin{tabular}{@{}ccccc@{}}
\toprule
& ResNet18 & ResNet34 & ResNet50 & ResNet101 \\ \midrule
FP & 69.75/89.07 & 73.31/91.42 & 76.13/92.86 & 77.37/93.55 \\
Tern & 66.01/86.78 & 70.95/89.89 & 74.00/91.77 & 75.63/92.49\\
Gap & -3.74/-2.29 & -2.36/-1.53 & -2.13/-1.09 & -1.74/-1.06 \\ \bottomrule
\end{tabular}}
\end{table}

Moreover, we conduct a series of deeper residual networks (ResNet-34, 50, 101) in addition to the ResNet-18. As listed in \cref{table_deepresnet_imagenet}, the accuracy gap between a ternarized model and its full-precision counterpart is reduced when network goes deeper. Such observation is in consistent with the experiments on the CIFAR-10 dataset.

\subsubsection{Improve accuracy on ResNet with REL} 

We introduce the Residual Expansion Layer (REL) technique to compensate the accuracy loss, which can still benefit from the weight ternarization in model size and computation reduction, and the multiplication elimination for convolution and fully-connected layers. In order to show that such residual expansion can effectively reduce the accuracy gap between ternarized model and full-precision baseline, we report the validation accuracy of expanded residual ResNet18 on ImageNet in \cref{table_resnet18_expand}. When expansion factor $T_{ex}$ is 2 and two thresholds $\beta=\{0.05, 0.1\}$ are used, we succeeded to further shrink the accuracy gap to -1.7\%/-1.03\%. When the $T_{ex}$ is increased to 4 ($\beta=\{0.05, 0.1, 0.15, 0.2\}$ ), the accuracy gap is almost negligible.

\begin{table}[h]
\centering
\caption{Validation accuracy (top1/top5 \%) of ResNet-18b on ImageNet with/without residual expansion layer (REL).}
\label{table_resnet18_expand}
\scalebox{1}{
\begin{tabular}{@{}cccccc@{}}
\toprule
     & 
\begin{tabular}[c]{@{}c@{}}First\\layer \end{tabular} & 
\begin{tabular}[c]{@{}c@{}}Last\\layer\end{tabular} & 
\begin{tabular}[c]{@{}c@{}}Accuracy\\ (top1/top5)\end{tabular} & \begin{tabular}[c]{@{}c@{}}Accuracy\\gap \end{tabular} & \begin{tabular}[c]{@{}c@{}}Comp.\\ rate\end{tabular} \\
\midrule
Full precision & FP & FP & 69.75/89.07 & -/- & 1$\times$ \\
\midrule
$T_{ex}$=1 & FP & FP   & 67.95/88.0 & -1.8/-1.0 & $\sim 16\times$ \\
$T_{ex}$=1 & Tern & Tern & 66.01/86.78 & -3.74/-2.29 & $\sim 16\times$ \\
\midrule
$T_{ex}$=2 & FP & FP & \textbf{69.33/89.68} & \textbf{-0.42/+0.61} & $\sim 8\times$ \\
$T_{ex}$=2 & Tern & Tern & 68.05/88.04 & -1.70/-1.03 & $\sim 8\times$ \\
\midrule
$T_{ex}$=4 & Tern & Tern & \textbf{69.44/88.91} & \textbf{-0.31/-0.16} & $\sim 4\times$ \\
\bottomrule
\end{tabular}}

\end{table}

\begin{figure}[h]
\centering
	\subfloat[\label{fig_accuracy_expand_ff_lf}]{%
		\includegraphics[width=0.24\textwidth]{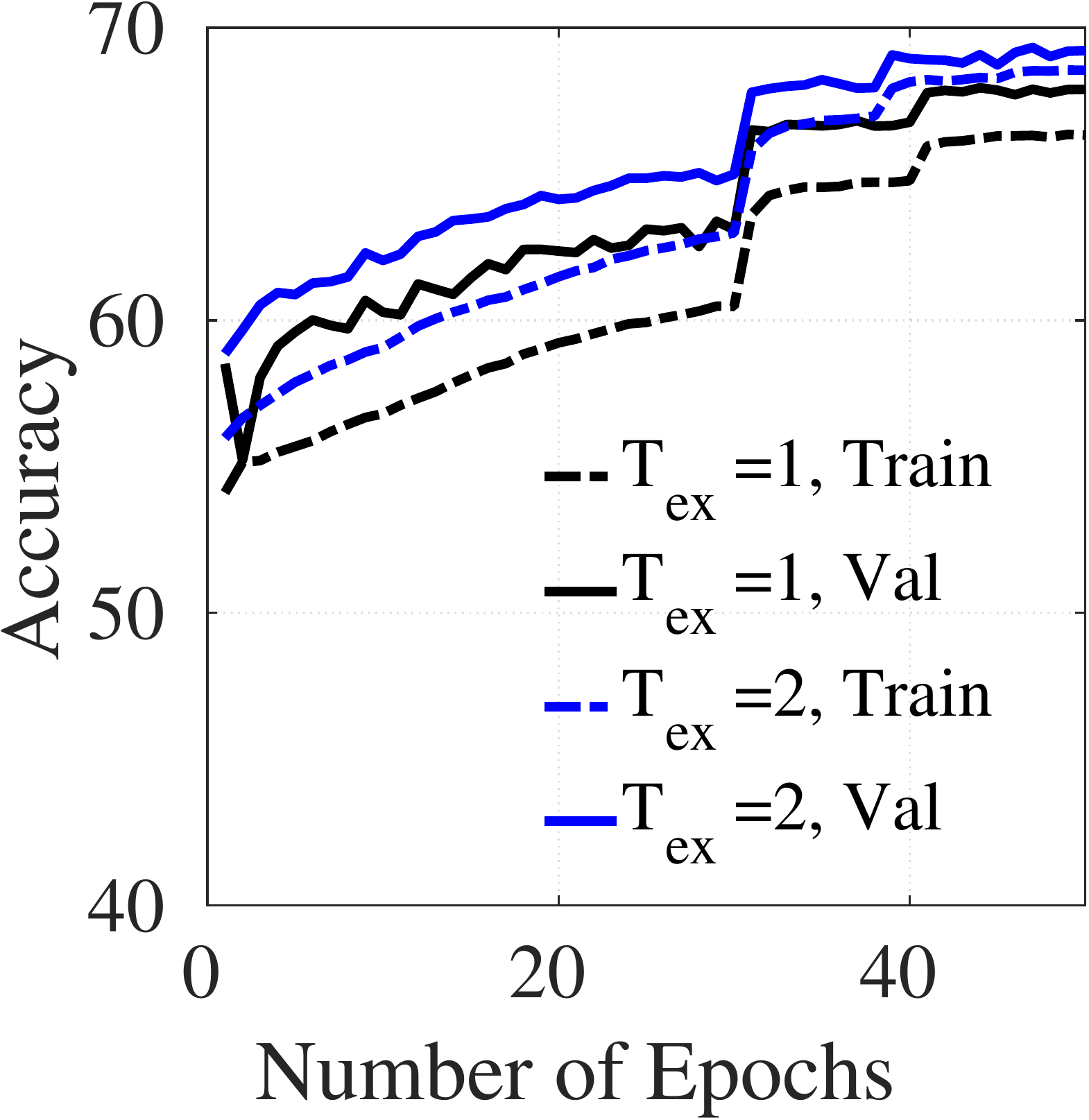}
	} 
	\subfloat[\label{fig_accuracy_expand_fq_lq}]{%
		\includegraphics[width=0.24\textwidth]{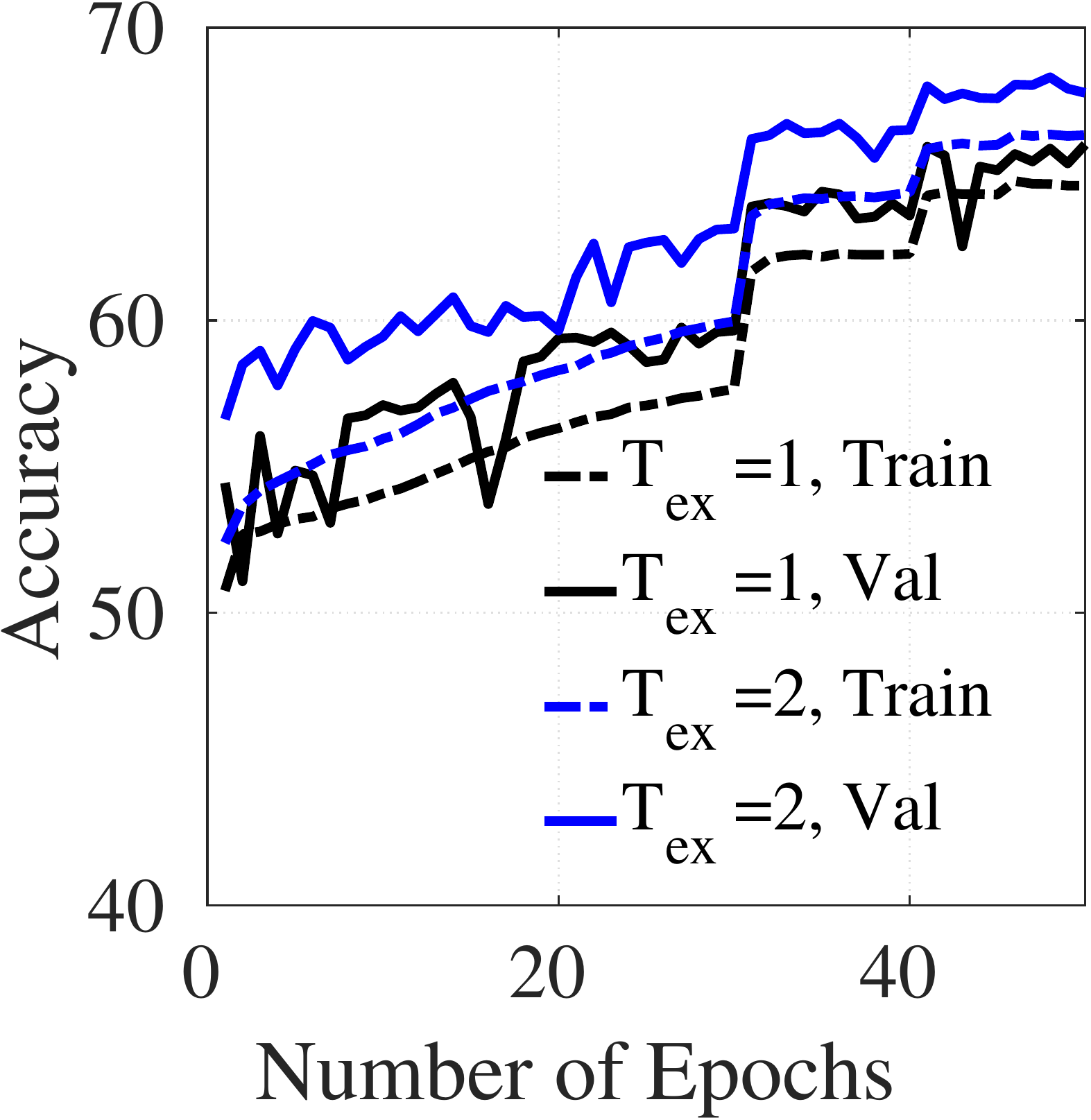}
	}
	\caption{The accuracy evolution curve of ternarized ResNet-18b on ImageNet with residual expansion. (a) First and last layer in full-precision; (b) First and last layer are ternarized} 
	\label{fig_accuracy_expand}
\end{figure}

The training and test accuracy evolution accuracy of ResNet18 with all the configurations listed in \cref{table_resnet18_expand} are plotted in \cref{fig_accuracy_expand}. Based on our observation, whether implementing ternarization of the first and last layer does have a significant effect on the training process and accuracy.
Ternarizing the first and last layer of ResNet18 (\cref{fig_accuracy_expand_fq_lq}) causes large validation accuracy fluctuation during training compared to the experiments that keep the first and last layer in full precision (\cref{fig_accuracy_expand_ff_lf}). As shown in \cref{fig_accuracy_expand_fq_lq}, the validation accuracy curve with residual expansion alleviate the fluctuation and smooth the curve.

\begin{figure}[ht]
\centering
	\subfloat[\label{fig_kernel_fp_before}Full-precision kernel before fine-tuning]{%
		\includegraphics[width=0.22\textwidth]{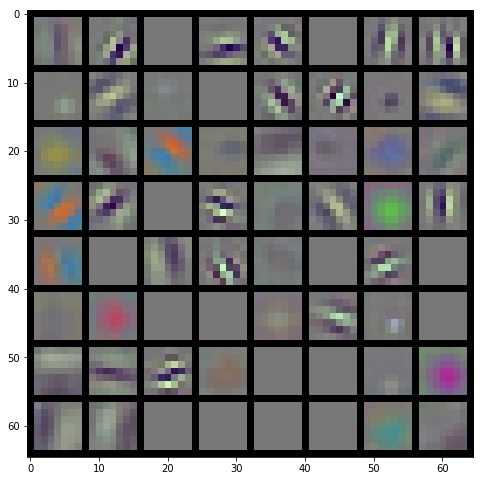}
	} \hspace*{\fill}
	\subfloat[\label{fig_kernel_fp_after}Full-precision kernel after fine-tuning]{%
		\includegraphics[width=0.22\textwidth]{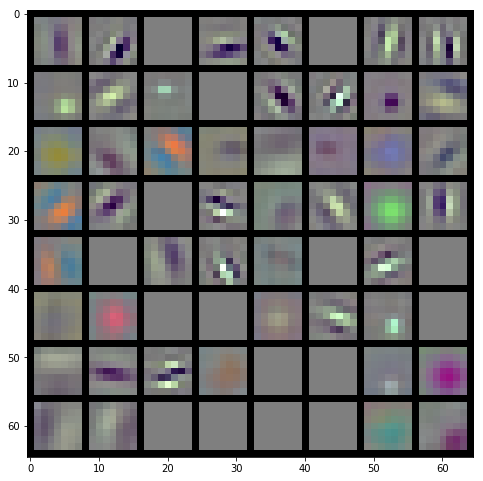}
	} \\
	\subfloat[\label{fig_kernel_tern1}Ternarized kernel with threshold factor $\beta=0.05$.]{%
		\includegraphics[width=0.22\textwidth]{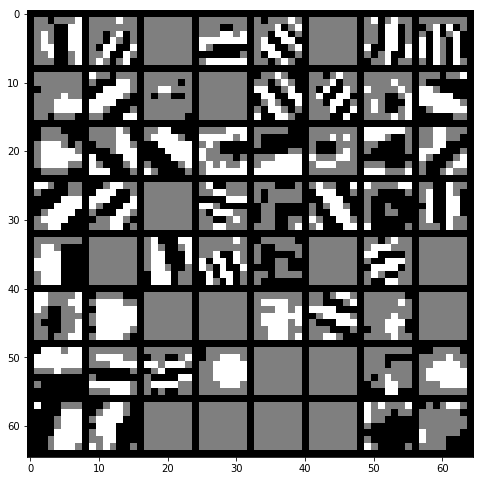}
	} \hspace*{\fill}
	\subfloat[\label{fig_kernel_tern2}Ternarized kernel with threshold factor $\beta=0.1$.]{%
		\includegraphics[width=0.22\textwidth]{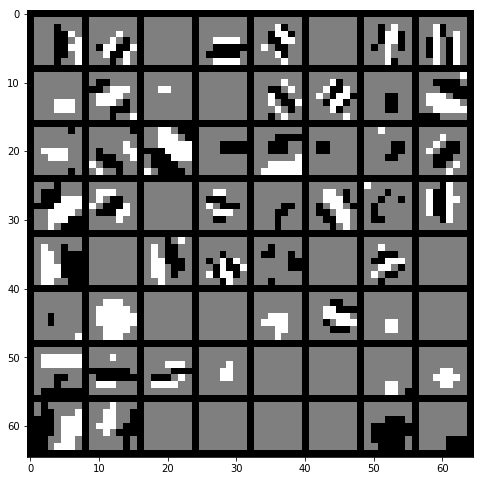}
	}	
	\caption{Kernel visualization for the first layer (channel 0) of ResNet18 before and after ternarization.} 
	\label{fig_kernel_visual}
\end{figure}

Beyond that, we further explore the effect of ternarization process of neural network with the assistance of kernel visualization. Since we also ternarize the first convolution layer of ResNet18, the input channel-0 is used as an example to study the weight transformation during weight ternarization as shown in \cref{fig_kernel_visual}. Kernels plotted in \cref{fig_kernel_fp_before} are from the full-precision pretrained model, which are taken as the initial weights. After fine-tuning the model for weight ternarization, the full-precision weights (i.e., the full precision weight is kept for back-propagation) shown in \cref{fig_kernel_fp_after} still keep the similar patterns as in the pretrained model. During inference in validation stage, kernels in \cref{fig_kernel_fp_after} are ternarized with two different threshold factors, $\beta=0.05$ and $\beta=0.1$ for the original layer \textit{Conv} and its REL \textit{Conv\textsubscript{r}} respectively. As seen in \cref{fig_kernel_tern1} and \cref{fig_kernel_tern2}, both ternarized kernels, to some extent, preserve the features from full precision model. Ternarized kernels with larger threshold factor only keep the connection with higher weight intensity. Thus, the residual expansion layer with higher threshold factor will lead to higher layer sparsity, and only a very small portion weights (i.e., non-zero value) will use the computing resources to perform network inference.

\section{Ablation study and discussion}
\subsection{Ablation study}
\begin{table}[h]
\centering
    \caption{Validation accuracy of ResNet-18b on ImageNet.}
\label{table_resnet_ablation}
\scalebox{1}{
\begin{tabular}{@{}ccc@{}}
\toprule
     & Accuracy & Comp. rate\\ \midrule
Baseline: Full precision  & 69.75/89.07 & 1$\times$ \\
\midrule
TW &  57.48/81.15 & $\sim$16$\times$ \\
TW+ICS ($\beta=0.05$) & 65.08/86.06 & $\sim$16$\times$ \\
TW+ICS ($\beta=0.1$) & 64.78/86.06 & $\sim$16$\times$ \\
TW+ICS ($\beta=0.2$) & 57.09/81.01 & $\sim$16$\times$ \\
TW+FT  & 64.94/86.13 & $\sim$16$\times$ \\
TW+ICS+FT ($\beta=0.05$) & 66.01/86.78 & $\sim$16$\times$ \\
TW+ICS+FT+REL ($\beta=\{0.05,0.1\}$)& 68.05/88.04 & $\sim$8$\times$ \\
\bottomrule
\end{tabular}}

\end{table}

In order to show the effectiveness of the proposed methods, we present a fair ablation study (\cref{table_resnet_ablation}) that isolates multiple factors which are listed as follow: 
1) TW: directly training the Ternarized Weight (TW) from scratch without Iteratively Calculated Scaling factors (ICS); 
2) TW+ICS: Training the ternarized network from scratch with the varying scaling factor ($\beta$=0.05, 0.1 and 0.2); 
3) TW+FT: fine tuning (FT) the ternarized weight. It shows that using ICS and training from scratch is close to the performance of fine-tuning the model without ICS. 4) TW+ICS+FT: Fine-tuning the the model with ICS further improves the accuracy. 5) TW+ICS+FT+REL: incorporating our proposed REL techniques leads to almost no accuracy loss.

\subsection{Deep CNN density examination}
\label{sec_density_exam}

\begin{table}[h]
\centering
\caption{Density variation of first and last layer before/after fine-tuning}
\label{table_sparsity}
\scalebox{1}{
\begin{tabular}{@{}cccccc@{}}
\toprule
 & \multicolumn{2}{c}{First layer} & \multicolumn{2}{c}{Last layer} & \multirow{2}{*}{\begin{tabular}[c]{@{}c@{}}Average\\ density\end{tabular}} \\ \cmidrule(r){1-5}
 & Before & After & Before & After &  \\ \midrule
AlexNet & 36.3\% & 53.6\% & 62.8\% & 67.5\% & 41.1\% \\ \midrule
ResNet18 & 41.7\% & 52.2\% & 54.8\% & 68.3\% & 40.1\% \\ \bottomrule
\end{tabular}}
\end{table}

\begin{figure*}[h]
\centering
	\subfloat[\label{fig_density_alexnet}AlexNet]{%
		\includegraphics[width=0.37\textwidth]{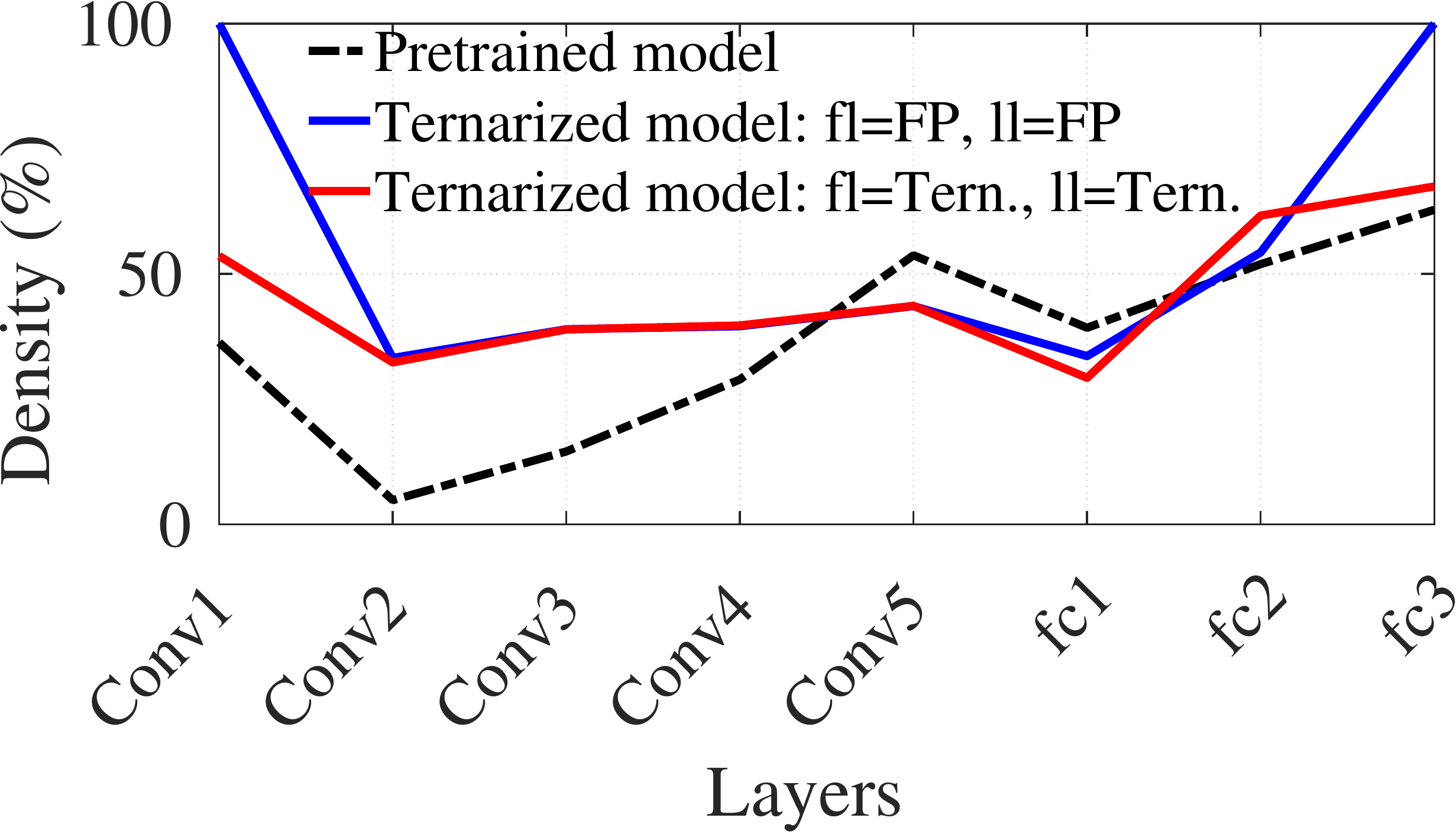}
	}\quad\quad
	\subfloat[\label{fig_density_resnet}ResNet-18]{%
		\includegraphics[width=0.4\textwidth]{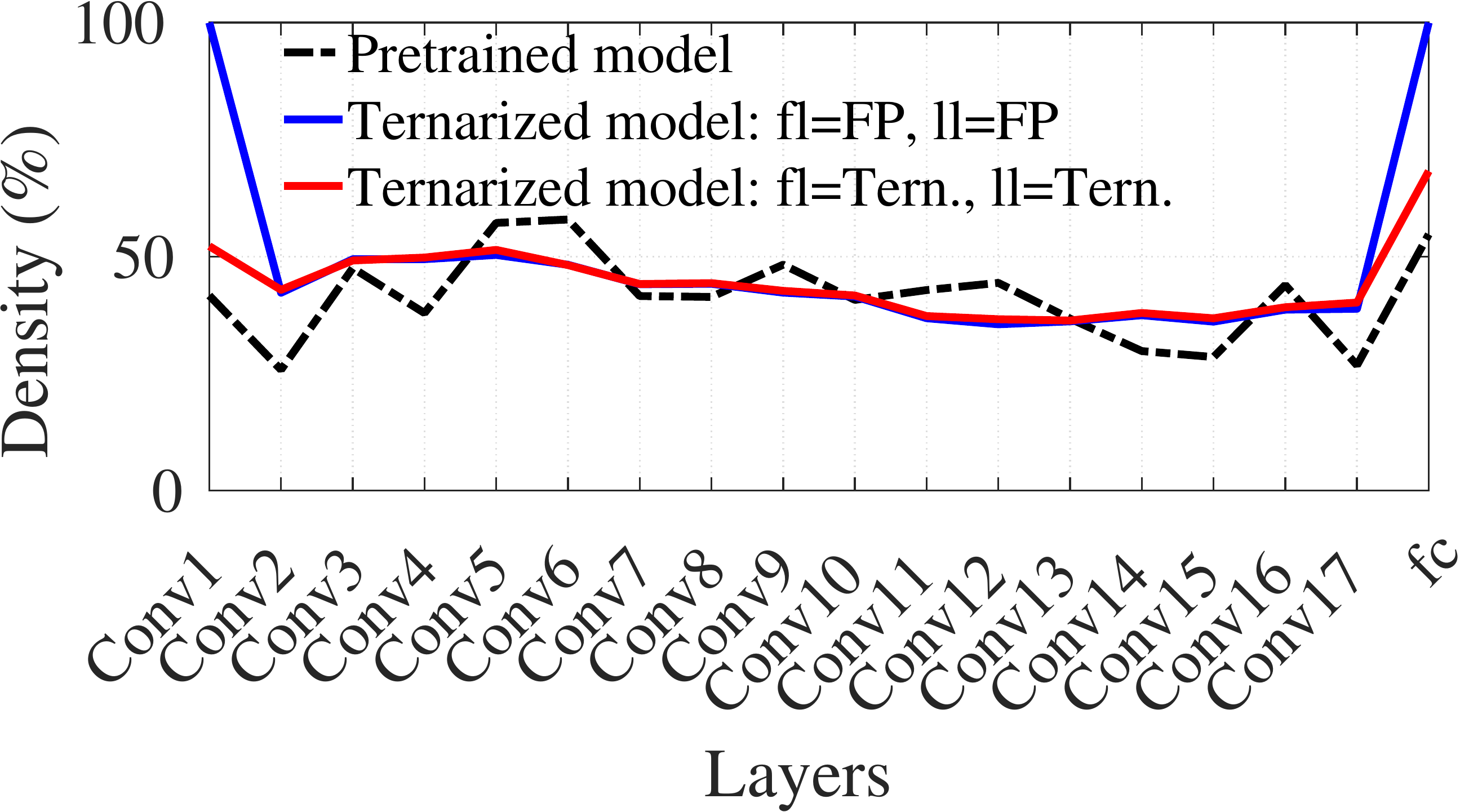}
	}
	\caption{Density distribution of convolution and fully-connected layers in (a) AlexNet and (b) ResNet-18 structure. The weight density of pretrained model is when the network just load the parameters from pretrained model without fine-tuning. Ternarized model: fl=FP, ll=FP refers the density distribution of the ternarized model with first and last layer in full precision. Ternarized model: fl=Tern., ll=Tern. denotes that the density distribution of the ternarized model with first and last layer in ternary representation.} 
	\label{fig_density}
\end{figure*}
Let us assume the trained sparsity (sparsity=1-density) of the designated layer is correlated to its redundancy.
\cref{fig_density} depicts the weight density distribution of all the convolution and fully-connected layers in AlexNet and ResNet for the cases of before and after the model ternarization. As shown in both \cref{fig_density_alexnet} and \cref{fig_density_resnet}, the weight density of first and last layer is raised to a higher level after ternarization in comparison to their initial density. Here the density on pretrained model refers to the neural network that initialized from the full-precision pretrained model, then we directly apply our ternarization method as introduced in \cref{sec_ternarization} without fine-tuning the weights. 
We calculate the average density of all the convolution and fully-connected layers, which are reported in \cref{table_sparsity}. It shows that after ternarization, the density of first and last layer is much higher than the average density and other layers. 
A potential method to further compress network model is to use the density as an indicator to choose different expansion factors for each each layer. 

\subsection{REL versus multi-bit weight }
\label{sec_multibit}

\begin{table}[h]
\centering
\caption{Inference accuracy of ResNet-18 on ImageNet with multi-bit \cite{zhou2016dorefa} weight or REL.}
\label{table_expand_network}
\scalebox{1}{
\begin{tabular}{@{}cccc@{}}
\toprule

& \multicolumn{1}{c}{Multi-bit  \cite{zhou2016dorefa}}  & \multicolumn{2}{c}{REL (this work)} \\ \midrule

Config. & 3-bit  & t\textsubscript{ex}=1 & t\textsubscript{ex}=2 \\

Accuracy & 65.70/86.82  & 66.01/86.78 & 68.05/88.04 \\

Comp. rate & 10.7$\times$   & 16$\times$ & 8$\times$ \\ \bottomrule

\end{tabular}}
\end{table}

In comparison to directly using multiple bits (multi-bit) weight, using REL is advantageous in terms of both accuracy and hardware implementations. First, REL shows better performance over the multi-bit method adopted from Dorefa-Net \cite{zhou2016dorefa} on ImageNet. Second, from the hardware perspective, REL preserves all network weights in the ternary format (i.e. -1, 0, +1), enabling the implementation of convolutions through sparse addition operations without multiplication. Increasing the number of expansion layers (t\textsubscript{ex}) merely requires more convolution kernels when one deploys the network in FPGA/ASIC. It is noteworthy that the REL with larger threshold factor owns higher sparsity, which requires less computational resources. On the contrary, when more bits are added in multi-bit, the convolution computation goes back to multiplication and addition operations, consuming more energy and chip area. Finally, both our work (\cref{sec_density_exam}) and the deep compression \cite{han2015deep} show that a deep neural network has different redundancies from layer to layer. If one employs the multi-bit scheme, we have to re-design the computing hardware with different bit-widths for the layers. Beyond that, since zero-value weights in REL with smaller threshold will be stay in zero in REL with larger threshold (as shown in \cref{fig_kernel_tern1} and \cref{fig_kernel_tern2}), the model compression rate can be further improved with proper weight sparse encoding method.

\subsection{Hardware resource utilization efficiency}

\begin{table}[h]
\centering
\caption{FPGA resource utilization (Kintex-7 XC7K480T) for AlexNet last layer}
\label{table_FPGA}
\scalebox{0.9}{
\begin{tabular}{@{}ccccccc@{}}
\toprule
\multirow{2}{*}{\begin{tabular}[c]{@{}c@{}}Weight\\ type\end{tabular}} & \multicolumn{2}{c}{MACs} & \multicolumn{2}{c}{LUT (slices)} & \multicolumn{2}{c}{DSP (slices)} \\ \cmidrule(l){2-7} 
 & Multiplier & Adder & required & available & required & available \\ \cmidrule(){1-7}
FP & 100 & 100 & 49600 & 74650 & 200 & 1920 \\
Tern. & 0 & 90 & 23490 & 74650 & 0 & 1920 \\ \bottomrule
\end{tabular}}
\end{table}
As we previously discussed in \cref{sec_alexnet}, fully ternarizing the first and last layer is greatly beneficial for developing the corresponding hardware accelerator on FPGA or ASIC, owing to the elimination of floating point convolution computing core. Here in this work, we analyze the deployment of the last fully connected layer of AlexNet into a Xilinx high-end FPGA (Kintex-7 XC7K480T) as an example. As reported by \cite{ehliar2014area}, floating-point adder cost 261 LUT slices on Kintex-7 , while the floating-point multiplier takes 235 LUT slices with additional 2 DSP48Es.  
The weights of last fully-connected layer in AlexNet is in the dimension of $4096\times1000$, which requires 1000 \textit{MACs} \cite{chen2017eyeriss} if performing the computation in the extreme parallel manner. However, due to the FPGA resource limitation, we consider the design scheme that allows 100 MACs for floating-point weight convolution, while the rest resources can only accommodate 90 MACs for ternary weight convolution. 
For the neural network with both ternary and full precision weights, balancing the hardware resources to build MAC cores for ternary and full precision weight becomes extremely difficult. Increasing the number of ternary weight MAC cores does improve the hardware resource utilization. However, it has to reduce the allocated floating point MACs, which at the same time reduces the computation parallelism and thus increases whole system latency. On the contrary, if we keep the current hardware resource allocation plan, the hardware utilization efficiency will be extremely low especially when the networks are going deeper. Then, it can be clearly seen that fully ternarizing the whole network is extremely important for algorithm deployment in resource limited embedded system although the theoretical model compression rate does not increase too much.

\section{Summary}

In this work, we have explicitly optimize current neural network ternarization scheme with techniques like statistical weight scaling and REL. On image classification task, a series of comprehensive experiments are conducted to show that our method can achieve the state-of-the-art accuracy on both small dataset like CIFAR-10, and large dataset like ImageNet. Beyond that, we examined the accuracy with or without the first and last layer ternarization. Owing to that the statistical weight scaling are used through the entire network, our ternarized network does not encounter serious accuracy degradation even with ternarized weight for first and last layer.
Our future work will focus on the automated network compression based on the techniques used in this work, with additional function of trainable/searchable threshold and layer-wise expansion factors.

{\small
\bibliographystyle{ieee}
\bibliography{egbib}
}

\end{document}